\documentclass{article}

\usepackage{arxiv}

\usepackage[utf8]{inputenc} 
\usepackage[T1]{fontenc}    
\usepackage{hyperref}       
\usepackage{url}            
\usepackage{booktabs}       
\usepackage{amsfonts}       
\usepackage{nicefrac}       
\usepackage{microtype}      
\usepackage{lipsum}
\usepackage{graphicx}

\usepackage{bm}
\usepackage{multirow}
\usepackage{adjustbox}

\title{STADEE: STAtistics-based DEEp Detection of Machine Generated Text \thanks{The Version of Record of this contribution is published in International Conference on Intelligent Computing 2023 and is available online at  \url{https://doi.org/10.1007/978-981-99-4752-2_60}}}

\author{
 Zheng Chen, Huming Liu \\
  School of Information and Software Engineering\\
  University of Electronic Science and Technology of China\\
  Chengdu, China 610054 \\
  \texttt{zchen@uestc.edu.cn}}

\begin{document}
\maketitle
\begin{abstract}
The emergence of large-scale pre-trained language models (PLMs), such as ChatGPT, creates opportunities for malicious actors to disseminate disinformation, necessitating the development of automated techniques for detecting machine-generated content. However, current approaches, which predominantly rely on fine-tuning a PLM, face difficulties in identifying text beyond the scope of the detector's training corpus. This is a typical situation in practical applications, as it is impossible for the training corpus to encompass every conceivable disinformation domain. To overcome these limitations, we introduce STADEE, a \textbf{STA}tistics-based \textbf{DEE}p detection method that integrates essential statistical features of text with a sequence-based deep classifier. We utilize various statistical features, such as the probability, rank, cumulative probability of each token, as well as the information entropy of the distribution at each position. Cumulative probability is especially significant, as it is explicitly designed for nucleus sampling, the most prevalent text generation algorithm currently. To assess the efficacy of our proposed technique, we employ and develop three distinct datasets covering various domains and models: HC3-Chinese, ChatGPT-CNews, and CPM-CNews. Based on these datasets, we establish three separate experimental configurations\textemdash namely, in-domain, out-of-domain, and in-the-wild\textemdash to evaluate the generalizability of our detectors. Experimental outcomes reveal that STADEE achieves an F1 score of 87.05\% in the in-domain setting, a 9.28\% improvement over conventional statistical methods. Furthermore, in both the out-of-domain and in-the-wild settings, STADEE not only surpasses traditional statistical methods but also demonstrates a 5.5\% enhancement compared to fine-tuned PLMs. These findings underscore the generalizability of our STADEE in detecting machine-generated text. 

\end{abstract}


\section{Introduction}
In recent years, there have been notable advancements in the field of natural language generation, particularly with the development of large-scale PLMs like ChatGPT \cite{chatgpt} and GPT-4 \cite{gpt4}. The texts produced by these models are of such exceptional quality that it can be challenging for humans to discern them from those written by people. In fact, according to a technical report by OpenAI, the majority of texts generated by GPT-2 were already indistinguishable from those written by humans \cite{solaiman2019release}. These PLMs have a broad range of applications, including story \cite{storygen} and dialogue generation \cite{dialoggen}, as well as code writing \cite{codegen}. Nonetheless, they can also be easily exploited by malicious actors to fabricate fake news \cite{fakenews1,fakenews2,fakenews3} and comments \cite{fakereview} for personal profit or political interference, thereby posing a significant threat to society. Therefore, it is imperative to explore automatic methods for detecting machine-generated text to identify disinformation and mitigate the likelihood of abuse \cite{survey}.

Detecting machine-generated text is commonly viewed as a binary classification task, leading many researchers to employ the widely-used text classification method of fine-tuning a Pre-trained Language Model (PLM) to identify machine-generated text \cite{fakenews1,solaiman2019release,fakenews3}. While this approach can achieve high performance on test data with a similar distribution to the training data, it often falters in real-world applications. In these situations, the generative model, method, and hyperparameters employed by potential adversaries are typically unknown, causing discrepancies between the test and training data distributions. As a result, the recognition rate for this method can decrease significantly, demonstrating poor generalization \cite{stiff}. Our observations indicate that this is primarily due to the method's heavy reliance on specific locally-preserved semantic and structural information in the training data, such as certain phrases and expressions, rendering it ineffective when test data is altered.

Alternatively, another class of detection method is based on statistical features, which posits that existing PLMs must employ various probability-corrected decoding algorithms to generate coherent text, ultimately altering the statistical distribution of the generated text. This method leverages a PLM to compute statistical features, such as rank, for each token in the text sequence. Subsequently, these tokens are grouped into distinct buckets based on their rank, with the number of tokens in each bucket serving as input. Detection can be performed through human observation or by training a logistic regression classifier \cite{GLTR}. By extracting statistical features, this method exhibits significantly better generalizability than fine-tuning the PLM. Nonetheless, it faces challenges in effectively combining different types of statistical features and may lose the text's sequence information, resulting in a currently low recognition rate.

In order to address the critical limitations of existing methods in terms of generalizability and recognition rate, we introduce a novel detection approach called \textbf{STA}tistics-based \textbf{DEE}p detection (STADEE). This method merges the superior generalization of statistical features with the robust automatic modeling capabilities of neural networks. In STADEE, a Pretrained Language Model (PLM) extracts various statistical features from the original text, encompassing well-known features such as probability, rank, information entropy, and a newly designed feature called cumulative probability, specifically designed for nucleus sampling \cite{topp}, a prevalent text generation sampling algorithm. Additionally, we transform the rank using the following equation: $r^{\prime} \leftarrow \log _{10} r$, since the extensive vocabulary of the PLM may lead to a long-tail effect, where numerous low-probability tokens exist at the tail of the distribution, but they exhibit significant differences in rank. These four features are combined into a tensor of shape $ [sequence\_length, 4] $, preserving the original sequence information of the text. Subsequently, this tensor is input into a 3-layer Transformer-Encoder \cite{attention} model to determine whether the text is machine-generated.

Generalizability is a crucial factor in assessing the effectiveness of a detector for practical applications. The significance of generalizability stems from the unpredictability of how malicious actors generate text, including their choice of pre-trained language models, fine-tuning corpora, prompts during generation, and decoding algorithm hyperparameters. Unfortunately, this field has seen relatively few studies and suffers from a lack of standardized datasets. Consequently, we not only utilize the publicly available HC3-Chinese dataset \cite{HC3} but also construct two additional datasets, ChatGPT-CNews and CPM-CNews, with distinct domains and language models. Employing these datasets, we devise three sets of detection experiments, encompassing in-domain, out-of-domain, and in-the-wild configurations. Our aim is to thoroughly assess the recognition rate and generalizability of the detectors. Such well-established experiments can more accurately gauge a detector's practical applicability in real-world scenarios.

The experimental results reveal that STADEE attains a commendable recognition rate in the in-domain experiment, with an F1 score of 87.05\%, which is 9.28\% higher than the traditional statistics-based method, GLTR. Furthermore, STADEE demonstrates exceptional generalizability, exhibiting an enhancement of more than 5.5\% in F1 score when compared to the method of fine-tuning a PLM in both out-of-domain (HC3-mix $\rightarrow$ ChatGPT-CNews) and in-the-wild (CPM-CNews $\rightarrow$ HC3-all) experiments. These findings suggest that STADEE is a more effective approach in practical applications.

To summarize, the key contributions of this research include:

1. We introduce STADEE, a \textbf{STA}tistics-based \textbf{DEE}p Detection method that combines statistical text features with sequence-based deep classifier to effectively discern machine-generated text.

2. Our method utilizes probability, rank, cumulative probability, and information entropy as statistical features to detect machine-generated text. These features are generated by a third-party language model, which means that our approach a truly black-box method.

3. We assemble and create three distinct datasets encompassing diverse domains and models, and establish three separate experimental configurations to evaluate the generalizability of machine-generated text detectors. . The data and models are all publicly available at \url{https://github.com/HMgithub111/STADEE}.

\section{Preliminaries and Related Works}

\subsection{Pre-trained Language Models and Text Generation}

\subsubsection{Pre-trained Language Models.}
Autoregressive language models are commonly used for modeling the probability of text sequences in the field of natural language generation. The probability of a given text is determined by the product of conditional probabilities for each token in the sequence, as illustrated in Equation (1), where $\mathbf{x}=x_1, x_2, \ldots, x_n$ represents a text sequence.
\begin{equation}
P(\mathbf{x})=\prod_{i=1}^n P\left(x_i \mid x_1, \ldots, x_{i-1}\right)
\end{equation}

Most contemporary PLMs are built on the Transformer encoder or decoder architecture \cite{attention} and are pre-trained on extensive corpora in a self-supervised manner. The primary objective during training is to maximize the $P(\mathbf{x})$. Subsequently, these models can be utilized directly or fine-tuned for specific downstream tasks. Some notable PLMs include RoBERTa \cite{roberta}, ChatGPT, GPT-4, among others.

\subsubsection{Text Generation and Decoding Algorithms.}
Text generation typically commences with a starting token or a prefix as the input condition. A pre-trained language model (PLM) is then employed to calculate the probabilities $\bm{p} \in R^{|V|}$ ($V$ is the vocabulary) for all tokens to be appeared in the subsequent position. Then, a decoding algorithm is employed to finally select a single token from the probability distribution.
This process iterates until an end token is generated or the generated text reaches the desired length.
Hence, the decoding algorithm plays a crucial role in text generation, on par with the significance of the PLM.

Decoding algorithms can be categorized into to classes: deterministic decoding and random sampling.
For open-ended text generation, random sampling is predominantly employed. This approach selects an output token from the vocabulary randomly, according to the probability distribution $\bm{p}$. Due to the long-tail effect of the distribution $\bm{p}$, pure random sampling may lead to the selection of numerous low-probability tokens, consequently affecting the quality of the generated text. As a result, various random sampling algorithms have been designed to enhance the sampling of high-probability tokens. For instance, temperature sampling divides the scores $\bm{s}$ by a temperature coefficient $T$, followed by the recalculation of the probability distribution $\bm{p}^{\prime}$. When $T$ is less than 1, this method increases the sampling of high-probability tokens.
Top-k sampling truncates the probability distribution to include only the top k tokens, then renormalizes their probabilities to obtain $\bm{p}^{\prime}$ \cite{storygen}.
Nucleus sampling, on the other hand, truncates the probability distribution to the top tokens whose cumulative probability exceeds a given threshold and subsequently renormalizes to obtain $\bm{p}^{\prime}$ \cite{topp}.
Text generated by nucleus sampling closely resembles human writing, making it the preferred decoding algorithm for text generation. In this work, we also employ nucleus sampling to construct the dataset.

\subsection{Detection of Machine-generated Text}
\subsubsection{Fine-Tuning a PLM.}
Based on the prior knowledge learned from a large amount of unlabeled data by a PLM, this popular method takes the raw text as input and extracts the hidden state of specified token (such as [CLS] used by BERT \cite{bert}) to be sent to a linear layer for classification. 
For example, Zellers et al. \cite{fakenews1} released a text generative model called GROVER and fine-tuned the GROVER for detection. They found increasing the size of the detection model can improve the accuracy and
proposed both the long-tail effect of random sampling and sampling algorithms that restrict this effect can leave artifacts to be captured by the detector. 
Solaiman et al. \cite{solaiman2019release} fine-tuned the RoBERTa model for detection and found that a detection model trained on the text generated by nucleus sampling can transfer well to detect text generated by other sampling algorithms and that a detection model trained on the text generated by large models can transfer well to detect text generated by small models. 
Subsequently, Uchendu et al. \cite{fakenews3} also verified that fine-tuning the RoBERTa model performed better than some existing detection models. 

However, the generalization of fine-tuning-based detection methods seems to be poor. Stiff et al. \cite{stiff} found this method can hardly be transferred directly to detect text generated by different generative models, sampling algorithms, or domains. But they did not provide a solution.

\subsubsection{Using the Statistical Features Extracted by a PLM.}
Generative models generate texts based on the probability distribution learned from the training corpus. To ensure that the generated text has higher readability, the model usually samples from the head of the probability distribution resulting in the lack of rare tokens compared to human-written text, which provides a way for detection. Gehrmann et al. \cite{GLTR} open-sourced a visualization detection tool called GLTR, which uses a PLM to calculate the probability, rank, and information entropy for each token in a sentence and then performs some statistical analysis. Furthermore, they divided all tokens of a sentence into four buckets according to the rank (1-10, 11-100, 101-1000, and 1000+) and further counted the number of tokens in each bucket as the input feature to train a logistic regression classifier. 
Later, Ippolito et al. \cite{ippolito} found that setting 50 buckets of rank can achieve higher performance in their experiments. 

Alternatively, there are some zero-shot methods. For example, Solaiman et al. \cite{solaiman2019release}
used a PLM to calculate the average log probability of a text and classified it based on a threshold. And DetectGPT detected machine-generated text by estimating their local log probability curvature \cite{Detectgpt}. In addition, there are also some methods based on unigram and bigram TF-IDF features or the bag-of-words feature \cite{solaiman2019release}.

\section{Machine-Generated Text Detection}

\begin{figure}
	\centering
    \includegraphics[width=0.66\textwidth]{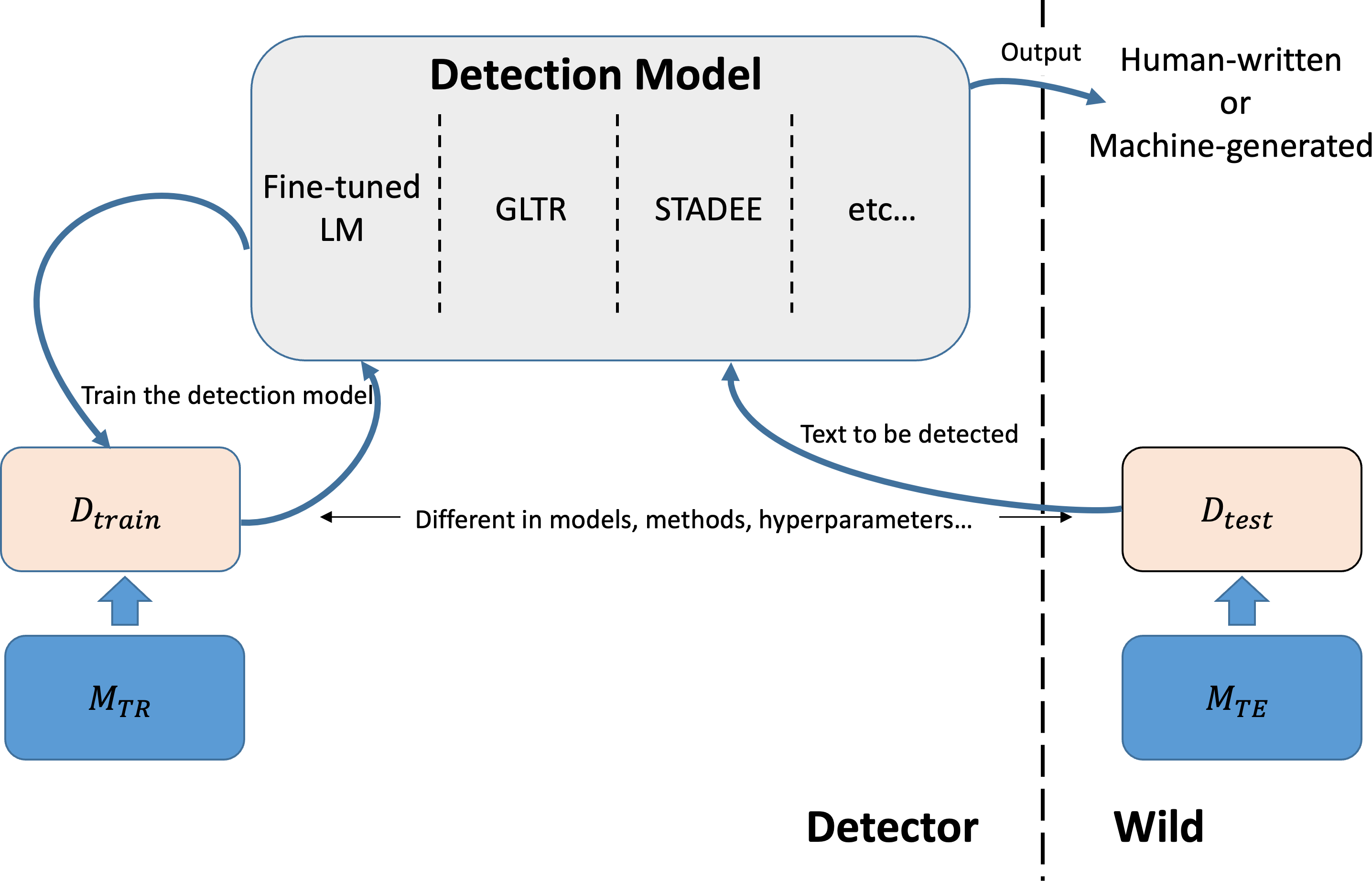}
    \caption{Common practical application scenarios for machine-generated text detection. Texts requiring detection often stem from diverse generative models, methods, and hyperparameters, which may differ significantly from those utilized in the training data. Consequently, it is essential for detectors to exhibit robust generalizability.} \label{fig1}
\end{figure}

\subsection{Task Definition}
Our study primarily focuses on detecting machine-generated text in supervised settings, which can be categorized as a binary classification task. More specifically, we aim to construct a model $f$: $f\left(x_i\right) \rightarrow y_i$, where $x_i$ is a piece of text, and $y_i \in\{0,1\}$ represents its corresponding label  (with label 0 indicating human-written text and label 1 denoting machine-generated text). 

In most supervised classification tasks, the test set $D_{test}$ and the training set $D_{train}$ are independent and identically distributed, allowing a well-trained model $f$ to perform equally well on both sets. However, in the case of machine-generated text detection, the distribution of $D_{test}$ often deviates from that of $D_{train}$. For example, $D_{test}$ may arise from diverse generative tasks (such as question-answering or text-completion), or entirely different generative models. Furthermore, human-written text may also vary in terms of styles and themes. Therefore, to ensure applicability across various scenarios, the detection model should possess strong generalizability.

To address these requirements, we have designed three experimental setups with progressively increasing difficulty to evaluate STADEE and other detection methods:

\begin{enumerate}
  \item \textbf{In-domain:} Both $D_{test}$ and $D_{train}$ are independent and identically distributed, as seen in most traditional classification tasks. Specifically, $D_{test}$ and $D_{train}$ are randomly divided after being generated using the same model, task, and identical parameters.

  \item \textbf{Out-of-domain:} $D_{test}$ and $D_{train}$ share the same generative model but employ different generative conditions. These conditions may vary in terms of the generation task, topic domain, or hyperparameters.

  \item \textbf{In-the-wild:} $D_{test}$ and $D_{train}$ utilize different generative models, which means the detector has no knowledge of how the text was generated. This setup aims to simulate real-world scenarios where the detection method faces previously unseen generation techniques.
\end{enumerate}

For the sake of clarity, we define the PLMs used in this study as follows:
\begin{itemize}
  \item $M_{TR}$: The model employed for constructing the training dataset.
  \item $M_{TE}$: The model used by malicious actors to generate text, which is typically distinct from $M_{TR}$.
  \item $M_E$: A critical component of the detection model, responsible for extracting statistical features from the text under examination. 
\end{itemize}
A comprehensive overview of the common practival application scenarios for machine-generated text detection task is depicted in Fig.~\ref{fig1}.

\subsection{The Limitations of Existing Detection Methods}

\subsubsection{Fine-Tuning a PLM.}

Due to remarkable memory abilities, large-scale PLMs tend to fixate to much on the local semantic features within the train set that are derived from the generative methods or preferences of the generative model. As a result, the detectors usually become entrenched in these fine-grained, localized features. When the semantic features in the test set are changed significantly, it becomes challenging to make accurate judgments.

\subsubsection{Using the Statistical Features Extracted by a PLM.}
To prevent overfitting in local semantic features, it is beneficial to extract statistical features such as probability and rank.
However, the current method GLTR solely relies on the partition of rank buckets. 
Determining an optimal partition and integrating multiple features can be challenging, while the loss of sequence information results in poor performance.
To address these issues, we propose the use of additional statistical features and a sequence neural network. 
Our aim is to automatically construct a more precise detector through an end-to-end approach that can distinguish subtle differences between human-written and machine-generated text.

\subsection{STADEE: STAtistics-based DEEp Detection}

\begin{figure}
    \includegraphics[width=\textwidth]{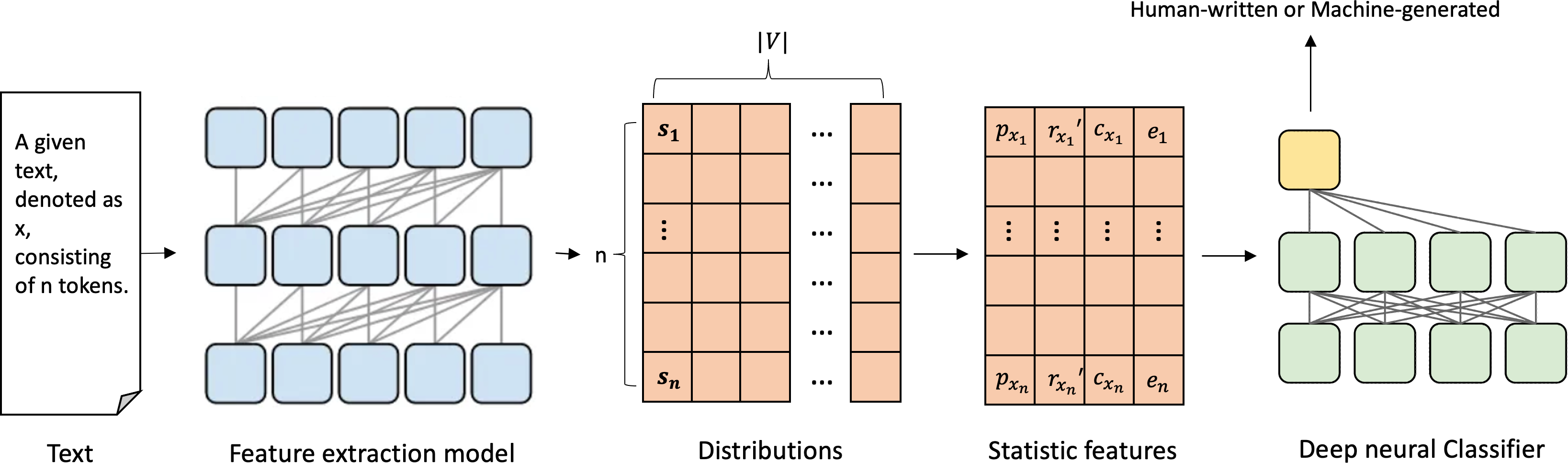}
    \caption{An Overview of STADEE: STAtistics-based DEEp detection. The system comprises two components: a feature extraction model $M_E$ and a sequence-based deep classifier. The $M_E$ extracts statistical features from the input text, after which the deep classifier makes a decision based on these statistical features.} \label{fig2}
\end{figure}

Statistical features serve as high-level attributes derived from text through the application of a Pre-trained Language Model (PLM). In the process, certain local semantic features that may hinder classification are discarded, thereby endowing statistical features with robust generalizability. Nonetheless, the complexity and diversity of these features pose a significant challenge when it comes to fully utilizing them.

Fortunately, neural networks, as end-to-end models, possess potent learning and representational capabilities. Sequential statistical features can be fed directly into the neural network, which could capable of automatically distinguishing between human-written and machine-generated text. Motivated by these insights, we propose a detection model named STADEE (refer to Fig.~\ref{fig2}), which is comprised of two components: a feature extraction model $M_E$ and a sequence-based neural classifier.

\subsubsection{Feature Extraction Model $M_E$.}

A PLM is used as $M_E$ to calculate statistical features for the text. 
Let $V=\left\{v_k\right\}_{k=1}^{|V|}$ denote the vocabulary of $M_E$, and let $\mathbf{x}=x_1, x_2, \ldots, x_n$ represent the text sequence of length n. At each position $x_i (1 \leq i \leq n)$, the score of all tokens in $V$ are calculated by $M_E$, denoted as $\bm{s}=\left\{s\left(v_k \mid x_1, \ldots, x_{i-1}\right)\right\}_{k=1}^{|V|}$.
Next, the following statistical features are calculated based on $\bm{s}$:\\\\

\begin{itemize}
  \item Probability: $p_{x_i} \in[0,1]$ is the probability of $x_i$ with respect to all tokens in $V$, as shown in Equation(2).
\begin{equation}
    p_{x_i}=p\left(x_i \mid x_1, \ldots, x_{i-1}\right)=\frac{s\left(x_i \mid x_1, \ldots, x_{i-1}\right)}{\sum_{k=1}^{|V|} s\left(v_k \mid x_1, \ldots, x_{i-1}\right)}
\end{equation}

  \item Rank: $r_{x_i} \in[1,|V|]$ is the rank of $x_i$ relative to all tokens in $V$.

  \item Cumulative probability: $c_{x_i}$ is the cumulative probability of $x_i$, as shown in Equation(3).
\begin{equation}
    c_{x_i}=\sum_{p_{v_k}>p_{x_i}} p_{v_k}
\end{equation}

  \item Information entropy: At the current position $i$, $e_i$ is the information entropy of the probability distribution of all tokens in $V$, as shown in Equation(4).
\begin{equation}
    e_i=-\sum_{k=1}^{|V|} p_{v_k} \log p_{v_k}
\end{equation}

 \end{itemize}


From the standpoint of the $M_E$, entropy represents the degree of uncertainty when the model evaluates a particular position, which can be understood as the potential information content of a given token. Probability, on the other hand, indicates the actual information content of this token. Both cumulative probability and rank are influenced by the presence of other possible tokens at this position, signifying whether the token can be placed within a favorable subset of all tokens at that position. Decoding from a favorable subset of tokens is a common decoding strategy employed by sampling algorithms in the natural language generation tasks.

Moreover, given that $V$ contains tens of thousands of tokens, the distribution inevitably comprises a substantial portion of low-probability tokens. This occurrence, referred to as the long-tail effect, results in considerable variation in the rank of low-probability tokens. To mitigate this issue, we implement a logarithmic correction to the rank: $r_{x_i}^{\prime}=\log _{10} r_{x_i}$. As discussed in the subsequent section, this approach has proven effective. Consequently, the four statistical features mentioned previously can be computed from a text of length $n$ and then consolidated into a 2-dimensional tensor, represented as $\bm{feature} \in R^{n * 4}$. This tensor serves as input for the downstream classifier.

\subsubsection{Sequence-based Neural Classifier.}

There are numerous model options for sequence-based neural classifiers, including Transformer, InceptionTime \cite{Inceptiontime}, and LSTM \cite{LSTM}. Among these models, we have opted for a Transformer-Encoder as our classifier. Specifically, we implement a three-layer Transformer-Encoder with a hidden size of 128 and a dropout rate of 0.3, followed by a linear classification layer with an input size of n*128, an output size of 2, and a dropout rate of 0.9.

It is essential to note that Transformer-based classifiers do not always outperform InceptionTime and LSTM-based ones. In Section 5.3, we conduct a series of experiments to assess the impact of the classifier. Both InceptionTime and LSTM demonstrate their advantages over the Transformer under specific configurations. Furthermore, we do not rule out the possibility that deep models with more parameters could exhibit even better performance. However, this would require the collection of additional training data.

\section{Experiment}
\subsection{Datasets}
ChatGPT has demonstrated its impressive capabilities in various domains, such as question answering, authoring, and coding, attracting extensive attention from both the industry and academia. However, this has further raised the concern about the potential misuse of machine text generation. Therefore, we target ChatGPT in our experiment. 
Specificly, we employ a publicly available Chinese dataset HC3-Chinese which is developed by ChatGPT using the method of question-answering. This dataset covers seven distinct domains, including finance, medicine, psychology, and more. We extract a total of 39,614 human-written and machine-generated answers from it.

For the out-of-domain experiment, we construct a second dataset named ChatGPT-CNews, which also utilizes ChatGPT.  Specifically, we leverage the THUCNews corpus \cite{cnews} and take the first 20 tokens of each sample as a prompt for ChatGPT to continue generating. We employ the gpt-3.5-turbo API with a hyperparameter top\_p set to 0.9 for generating this dataset, which distinguishes it from the HC3-Chinese dataset, as the latter was created using automated testing tools to extract data from the ChatGPT website. 

For the in-the-wild experiment, we construct a third dataset named CPM-CNews. It is generated by using CPM$_{THUCNews}$, which is fine-tuned from the CPM\footnote{\url{https://huggingface.co/TsinghuaAI/CPM-Generate}}\cite{cpm} on 80,000 samples in THUCNews for 5 epochs.  In particular, we use CPM$_{THUCNews}$ to continue generating based on the first 20 tokens of each sample in THUCNews. The nucleus sampling with a hyperparameter of 0.9 is used and the samples that are not used for generations are used as human-written text.

The size and  construction details  of the aforementioned datasets are shown in the Table~\ref{tab1}. Some conspicuous features in text are removed, such as unifying commas to Chinese commas, deleting symbols such as `\textbackslash n' and `\textbackslash\textbackslash n'.

\begin{table}[ht]
    \begin{center}
        \caption{The construction details and the size of the datasets.}\label{tab1}
        \begin{tabular}{lllcc}
        \toprule
        \multirow{2}{*}{Dataset} & \multirow{2}{*}{Generation Model} & \multirow{2}{*}{Generation Task} & \multicolumn{2}{c}{Size} \\
        \cline{4-5}
                                 &                                   &                                    & Human         & Machine  \\
        \hline
              HC3-Chinese        & ChatGPT                           & Question-Answering                               & 22231         & 17383     \\
              ChatGPT-CNews      & ChatGPT                           & Text-Completion                         & 2000          & 2000       \\
              CPM-CNews          & CPM$_{THUCNews}$                  & Text-Completion                         & 40000         & 40000    \\
        \bottomrule          
        \end{tabular}
    \end{center}
\end{table}

By using these datasets, we can setup three experiment configurations, which are:

\begin{itemize}
  \item  In-domain: we utilize all of the text in HC3-Chinese. The texts are randomly shuffled
and divided into $D_{train}$ and $D_{test}$ with an 8:2 ratio.

  \item  out-of-domain: we employ HC3-Chinese as $D_{train}$ and ChatGPT-CNews as $D_{test}$.
 
  \item  In-the-wild: we employ CPM-CNews as $D_{train}$ and HC3-Chinese as $D_{test}$.
  
\end{itemize}

\subsection{Baselines and Implementation Details}

For the baseline based on fine-tuning a PLM, we utilize the RoBERTa model hfl/chinese-roberta-wwm-ext\footnote{\url{https://huggingface.co/hfl/chinese-roberta-wwm-ext}}.
For the detection model based on statistical features, such as GLTR and our STADEE, 
we use the IDEA-CCNL/Wenzhong-GPT2-110M\footnote{\url{https://huggingface.co/IDEA-CCNL/Wenzhong-GPT2-110M}}to extract features from the raw text.
All samples are then truncated to the first 150 tokens and samples with insufficient length are removed (which are also removed in the fine-tuning baseline to ensure consistency). 
These models are available in the Hugging Face library \cite{transformers}.

When fine-tuning the RoBERTa, we use a batch size of 48, a learning rate of $5 \times e^{-5}$, and fine-tune 2 epochs. For GLTR, we use a batch size of 128 and train 500 epochs. For our STADEE, we use a batch size of 512, a learning rate of $4 \times e^{-5}$, and train 100 epochs. The AdamW optimizer is utilized for all the above methods.

\subsection{Results}

Table~\ref{tab2} shown the F1 scores of different detectors, with results reported for both validation and test datasets to evaluate the generalizability of the detectors. The validation datasets are derived from the training dataset, using a 1:7 ratio. The experiment results are presented as ($F_{valid}$, $F_{test}$), where the first item $F_{valid}$ denotes the result on the validation data, and the second item $F_{test}$ denotes the result on the test data. The best result on the \textbf{test} data for each row is highlighted in bold.

In the \textbf{in-domain} configuration, the training and  test dataset are independent and identical distributed.
Hence, fine-tuning RoBERTa yields the best results, which is unsurprising since it trained over the raw text that containing numerous fine-grained low-level features unique to this type of data. However, such features also leads to a decrease in generalization on other distributed data. In this experiment configuration, our proposed method STADEE achieves an decent F1 score of 87.05\%, outperforming the traditional statistical method GLTR by 9.28\%.

In the \textbf{out-of-domain} configuration, the distribution of the test dataset differs from that of the training dataset.
The performance of fine-tuned RoBERTa significantly drops on the test data, which is lower than that of STADEE. In the experiment of HC3-mix$\rightarrow$ChatGPT-CNews, STADEE outperforms fine-tuning the RoBERTa by 5.58\% on the test data, while still maintaining a high recognition rate of 87.74\% on the validation data.

In  the \textbf{in-the-wild} configuration, the distribution of the test data differs significantly from that of the training data.
The results exhibit a similar pattern to the out-of-domain configuration, but more pronounced.

In summary, fine-tuned RoBERTa exhibits excellent performance on test datasets with the same distribution but suffers from poor generalizability. On the other hand, GLTR possesses better generalization capabilities but yields a lower recognition rate. STADEE consistently maintains a stable and relatively high recognition rate under various experimental configurations, indicating its superior generalizability. This attribute is crucial for practical applications in real-world scenarios.

\begin{table}[ht]
    \begin{center}
        \caption{The F1 scores (\%) of different detectors on each set of experiments are presented in the format of ($F_{valid}$,$F_{test}$). $F_{valid}$ denotes the F1 on the validation data, while $F_{test}$ represents the F1 on the test data. The Transformer serves as the downstream sequence classifier of STADEE. Only the best result on the test data is highlighted in bold.}\label{tab2}
        \begin{tabular}{llccc}
        \toprule
            &  Train Data $\rightarrow$ Test Data   &   RoBERTa     &   GLTR          &    STADEE    \\
        \midrule
        \textbf{In-domain}     &  HC3-all $\rightarrow$ HC3-all               & (97.96,\textbf{98.37}) &   (78.19,77.77) & (87.65,87.05)  \\
        \midrule
        \multirow{5}{*}{\textbf{Out-of-domain}}     &  HC3-finance $\rightarrow$ ChatGPT-CNews     &  (99.78,73.56)   &   (87.91,73.38)   &   (95.13,\textbf{82.24})  \\
                                                  &  HC3-medicine $\rightarrow$ ChatGPT-CNews    &  (100.0,81.73)   &   (80.59,68.22)   &   (83.65,\textbf{85.11})  \\
                                                  &  HC3-psychology $\rightarrow$ ChatGPT-CNews  &  (99.21,70.87)   &   (48.39,52.77)   &   (77.90,\textbf{80.41})  \\
                                                  &  HC3-law $\rightarrow$ ChatGPT-CNews         &  (100.0,66.95)   &   (89.43,66.67)   &   (89.26,\textbf{68.64})  \\
                                                  &  HC3-mix $\rightarrow$ ChatGPT-CNews         &  (99.64,80.77)   &   (73.76,68.48)   &   (87.74,\textbf{86.35})  \\
        \midrule
        \textbf{In-the-wild}   &  CPM-CNews $\rightarrow$ HC3-all             &  (96.81,76.46)   &   (65.30,78.05)   &   (83.95,\textbf{82.62})  \\
        \bottomrule
        \multicolumn{5}{l}{To demonstrate the generalizability of detectors, we provide the F1 scores for both the validation and test data,} \\
        \multicolumn{5}{l}{presented in the format of ($F_{valid}$,$F_{test}$).}
        \end{tabular}
    \end{center}
\end{table}

\section{Discussion}

\subsection{The Impact of Different Statistical Features on STADEE}

\begin{table}[ht]
    \begin{center}
        \caption{The F1 scores (\%) of STADEE on the test data of each experiment respectively using the single statistical feature and the combination of features. The best result for each row is marked in bold.}\label{tab4}
        \begin{tabular}{llccccc}
        \toprule
          &\multirow{2}{*}{Train Data $\rightarrow$ Test Data}   & \multicolumn{5}{c}{Features used by STADEE} \\
        \cline{3-7}
                                   &  &   $p$     &    $r^{\prime}$     &    $e$    &    $c$    &    $p,r^{\prime},e,c$  \\
        \hline
        \textbf{In-domain}  & HC3-all $\rightarrow$ HC3-all       &  82.80    &    82.77         &   78.71   &   84.62            &    \textbf{87.05}  \\
        \textbf{out-of-domain} & HC3-mix $\rightarrow$ ChatGPT-CNews &  78.59    &    79.92         &   67.96   &   \textbf{86.59}   &    86.35  \\
        \textbf{In-the-wild} & CPM-CNews $\rightarrow$ HC3-all     &  78.94    &    78.70         &   49.44   &   79.87            &    \textbf{82.62}   \\
        \bottomrule
        \end{tabular}
    \end{center}
\end{table}

To examine how various statistical features affect the recognition rate and generalizability of STADEE, we conduct experiments using the single statistical feature and the combination of features respectively. Also, the Transformer mentioned in section 3.3 is used as the downstream sequence classifier. The F1 scores on the \textbf{test} data of each experiment are presented in Table~\ref{tab4}. The in-domain configuration reflects the recognition rate of the detector under the same distribution data, while out-of-domain and in-the-wild configuration examine generalizability.

After reviewing the results, we have found that the cumulative probability feature outperforms all other individual features.
This suggests that the introduction of this feature in our study is both necessary and correct.
Additionally, the combination of features improves the performance of the detector except for a slight decrease in generalizability in \textbf{out-of-domain} configuration. Notably, in \textbf{in-the-wild} configuration, the single information entropy feature performs worse than random selection. However, the result of combining it with other features surpasses that of any single feature.
This outcome provides further evidence supporting the importance of utilizing multiple features.

\subsection{The Effectiveness of the Logarithmic Transformation on Rank}

\begin{table}[ht]
    \begin{center}
        \caption{The F1 scores (\%) of STADEE on the test data of \textbf{in-domain} configuration, utilizing both the original single rank and the rank after logarithmic transformation.}\label{tab3}
        \begin{tabular}{lcc}
        \toprule
          & \multicolumn{2}{c}{Features used by STADEE}  \\
        \cline{2-3}
          & $r$     &    $r^{\prime}=\log _{10} r$  \\
        \hline
        \textbf{In-domain}&   67.30   &    \textbf{82.77}   \\
        \bottomrule
        \end{tabular}
    \end{center}
\end{table}

To evaluate the efficacy of the logarithmic transformation on rank in STADEE, we conduct
experiments on \textbf{in-domain} configuration, utilizing both the original single rank and the rank after logarithmic transformation. The Transformer architecture mentioned in
section 3.3 is used as the downstream sequence classifier. The F1 scores on the \textbf{test} data are
shown in the Table~\ref{tab3}.

The result indicates that applying logarithmic transformation to the rank has increased the F1 score by 15.47\% as compared to the original rank. 
This implies that to a certain extent, this transformation
can correct the bias existing in the original rank of whether the token is sampled from the top.

\subsection{The Impact of Different Downstream Sequence Neural Networks on STADEE}

\begin{table}[ht]
    \begin{center}
        \caption{The F1 scores (\%) of STADEE on the test data of each experiment respectively using the different downstream sequence classifiers. The best result for each row is marked in bold.}\label{tab5}
        \begin{tabular}{llccc}
        \toprule
           &\multirow{2}{*}{Train Data $\rightarrow$ Test Data}    & \multicolumn{3}{c}{Downstream classifiers used by STADEE}\\
        \cline{3-5}
          &   &   InceptionTime     &    LSTM     &    Transformer    \\
        \hline
        \textbf{In-domain}  &HC3-all $\rightarrow$ HC3-all    &  86.38              &    85.78             &    \textbf{87.05}    \\
        \textbf{out-of-domain} &HC3-mix $\rightarrow$ ChatGPT-CNews &  86.83           &    \textbf{88.95}    &    86.35     \\
        \textbf{In-the-wild} &CPM-CNews $\rightarrow$ HC3-all     &  \textbf{84.15}  &    83.74             &    82.62   \\
        \bottomrule
        \end{tabular}
    \end{center}
\end{table}

In this section, we have experimented with several sequence classifiers, including InceptionTime based on 
CNN, LSTM based on RNN, and Transformer based on attention mechanism. The F1
scores on the \textbf{test} data of each experiment are shown in Table~\ref{tab5}.
The result indicates that the Transformer performs the best on the same distribution data, but other structures appear to have slightly better generalizability.

\section{Conclusion and Future Work}
In this paper, we have presented a novel detection model, STADEE, which integrates a statistical feature extractor with a sequence classifier for the identification of machine-generated text. The proposed model, STADEE, exhibits substantial improvement over traditional statistical methods and outperforms fine-tuned PLMs in terms of generalization capabilities, rendering it better suited for complex real-world scenarios.

Nonetheless, there remain several aspects of STADEE that warrant further improvement and exploration: a) Delving into more effective statistical features, as well as examining potential feature transformations and combinations. b) Incorporating deeper and more powerful downstream classifiers.

\bibliographystyle{unsrt}  
\bibliography{template}

\end{document}